%% file: neurips_2024.tex
\documentclass{article}

\usepackage[preprint]{neurips_2024}

\usepackage[utf8]{inputenc} 
\usepackage[T1]{fontenc}    
\usepackage{hyperref}       
\usepackage{url}            
\usepackage{booktabs}       
\usepackage{amsfonts}       
\usepackage{nicefrac}       
\usepackage{microtype}      
\usepackage{xcolor}         

\usepackage{algorithm}      

\usepackage{comment}
\usepackage{pifont}
\usepackage{marvosym}

\usepackage{csquotes}
\usepackage[font=itshape]{quoting}
\usepackage{colortbl}
\usepackage{xspace}
\usepackage{graphicx}
\usepackage{caption}
\usepackage{subcaption}
\usepackage{amsmath}
\usepackage[super]{nth}

\newcommand{\name}{\textsc{LLMSteer}\xspace}
\newcommand{\mypara}[1]{\vspace{0.1cm}\noindent{\bf {#1}:}~}

\title{\name: Improving Long-Context LLM Inference by Steering Attention on Reused Contexts}

\author{%
  Zhuohan Gu\thanks{Equal contribution.} \\
  University of Chicago\\
  \texttt{zhuohan@uchicago.edu} \\
  \And
  Jiayi Yao\footnotemark[1] \\
  University of Chicago \\
  \texttt{jiayi3@uchicago.edu} \\
  \AND
  Kuntai Du \\
  University of Chicago \\
  \texttt{kuntai@uchicago.edu} \\
  \And
  Junchen Jiang \\
  University of Chicago \\
  \texttt{junchenj@uchicago.edu} \\
}

\begin{document}

\maketitle

\input{sec_abstract}
\input{sec_intro}
\input{sec_background}
\input{sec_design}
\input{sec_eval}
\input{sec_future}
\input{sec_ending}

\bibliographystyle{plainnat}
\bibliography{citations}
\label{sec:ref}

\newpage
\appendix

\section{Appendix / supplemental material}

\subsection{Prefix prompts} \label{subsec:appendix:prefixes}
\begin{figure}[!htbp]
    \centering
    \includegraphics[width=1.0\columnwidth]{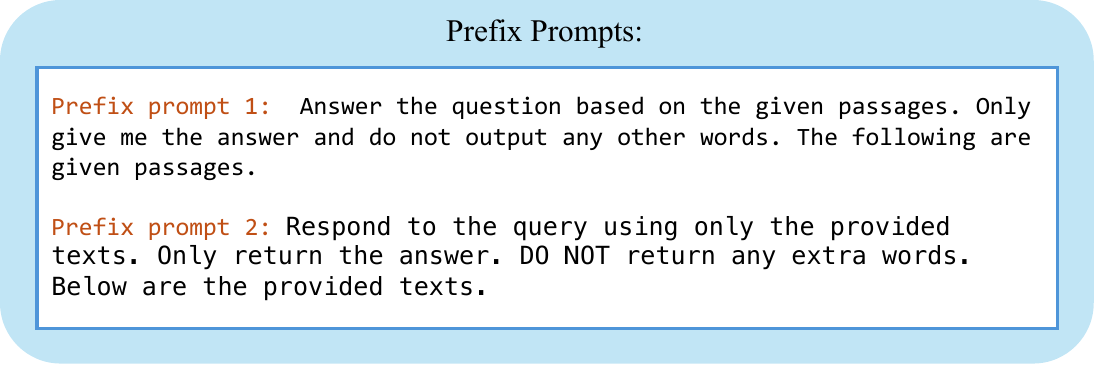}
    \vspace{-10pt}
    \caption{The two prefix prompts used in \name to generate different key-value caches. By processing the same context with these varied prefixes, \name encourages the model to interpret the context differently in each pass.}
    \label{fig:prefixes}
\end{figure}

\subsection{\name Algorithm} \label{subsec:appendix:algo}
\begin{algorithm}
\caption{\name}
\noindent\textbf{Input} A context $C$, a set of queries $\{q_1, q_2, \dots, q_n\}$, an LLM $\mathcal{L}$, prefix prompts $P_1, P_2$, and a scaling factor $\alpha$.\\
1: Contextual re-reading: $A_i = \text{Attention}_\mathcal{L}(P_i \oplus C)$ for $i \in \{1,2\}$;\\
2: Token selection: $T(l) = \bigcap_{i=1}^2 \text{TopK}(\sum_{h} A_i^{l,h}, k)$, where $l$ denotes the layer and $h$ denotes the attention head;\\
3: Steering: $A_{\text{steered}} = \mathbf{M} \odot A$, where $\mathbf{M}$ denotes a weighting matrix;\\
\textbf{Output} Modified attention weights for context $C$: $A_{\text{steered}}$
\end{algorithm}

\end{document}

%% file: sec_abstract.tex
\begin{abstract}
    As large language models (LLMs) show impressive performance on complex tasks, they still struggle with longer contextual understanding and high computational costs. To balance efficiency and quality, we introduce \name, a fine-tuning-free framework that enhances LLMs through query-independent \textit{attention steering}. Tested on popular LLMs and datasets, \name narrows the performance gap with baselines by 65.9\% and reduces the runtime delay by up to 4.8× compared to recent attention steering methods.
\end{abstract}

%% file: sec_intro.tex
\section{Introduction}
\label{sec:intro}
\vspace{-2mm}
Large Language Models (LLMs) have demonstrated remarkable capabilities in complex tasks such as question answering, summarization, and reasoning (~\cite{llm-app-1, llm-app-2, llm-app-3}). To enhance their reliability, LLMs are often augmented with domain-specific or user-specific knowledge that extends beyond their inherent training data (~\cite{rag1, rag2, rag3}). However, incorporating these supplemental contexts, which can exceed thousands of tokens (~\cite{jin2024ragcache, gao2023ragsurvey}), presents two challenges: (1) models often struggle to comprehend \textit{long context} (e.g., lost-in-the-middle problem (~\cite{liu2023lostmiddlelanguagemodels, junqing2023never})) and (2) processing long context incurs substantial runtime costs (\cite{liu2024farewell, lin2024infinite, zhong2024distserve}).

Since the Key-Value (KV) cache of the same context text chunks is often reused multiple times (\cite{cachegen, yao2024cacheblend, jin2024ragcache}), many recent systems adopt prefix caching (\cite{jin2024ragcache, cachegen, qin2024mooncake}), which stores the KV caches for the frequently reused contexts such that LLMs no longer need to prefill these contexts repeatedly. However, the model persists in losing track of key information from the context as its KV pairs remain unchanged. \textbf{So, is there a way to simultaneously achieve high efficiency and high quality without fine-tuning models?}

We present \name, a pioneering, fine-tuning-free framework that efficiently improves the generation quality of models through \textit{post-hoc attention steering}. More specifically, \name reweights attention scores to steer the model's attention toward selected tokens, thereby improving its contextual understanding. Our insight comes from the fact that storing and reusing the KV cache offers an opportunity to modify the KV cache offline to improve the model's understanding regarding the context. Thus, all queries could benefit from this improvement as the process of modifying the KV cache of the preceding context is independent of the queries. \name enables LLMs to read the same context in different ways by leveraging the idea that each pass through the same context with different prefix prompts leads the model to generate distinct KV caches, which represent different understandings of the context.

We implemented \name on top of LLaMA-3.1-8b-Instruct (~\cite{dubey2024llama}), and compared \name with LLaMA-3.1-8b-Instruct, LLaMA-3.1-70b-Instruct (~\cite{dubey2024llama}) and the state-of-the-art attention steering baseline. \name demonstrates that by steering tokens that are consistently important across the model's understandings, it not only achieves a significant increase in F1 scores (e.g., 72.9 -> 82.0), but also delivers up to 4.8× faster compared to the baselines. \name not only reduces runtime costs but also improves generation quality. This paper is the first effort to (1) improve model generation quality without fine-tuning and (2) do so in a way that is compatible with prefix caching.

%% file: sec_background.tex
\section{Post-hoc attention steering}
\vspace{-2mm}


Attention steering didn't come out of nowhere. Methods like PASTA (~\cite{zhang2023tellmodelattendposthoc}) and AutoPASTA (~\cite{anonymous2024model}) introduce techniques to boost the generation quality of LLMs by guiding their attention to specific parts of the input text. In PASTA, users manually specify tokens in the input and steer the LLM's attention toward these highlighted tokens through reweighting. However, annotating relevant input spans by humans is not always feasible, especially for lengthy contexts and context-specific tasks.

AutoPASTA enhances PASTA by automating the process of identifying key contextual information through iterative prompting. More specifically, AutoPASTA directly prompts the LLM to generate the key sentence in the context based on the query, and then upweight the attention scores of tokens in the key sentence. Despite its potential in improving model's generation quality, AutoPASTA has several limitations. 

First, the iterative prompting approach necessitates two LLM calls, which increases runtime latency. Second, AutoPASTA is incompatible with prefix caching, a technique in which KV caches of the frequently used text chunks are stored and reused such that the prefill of the reused text chunks can be skipped. For example, Llama-8B (~\cite{dubey2024llama}) requires 2.04 seconds to process a 5000-token context using 2 A40 GPUs. However, if the KV cache of the text chunk is precomputed and stored, only the final user query needs to be prefilled, which only takes 0.039 seconds on the same hardware. Although prefix caching can still be applied to the initial call in AutoPASTA, the second call cannot benefit from prefix caching in that the attention of the context is steered in a query-dependent way. Consequently, the KV cache of the context will be altered and can no longer be reused across different queries.

%% file: sec_design.tex
\section{\name design}
\label{sec:design}
\vspace{-2mm}
We now present the design of \name, which not only reduces runtime costs but also improves inference quality. We begin with the basic insight, followed by how \name selects and steers tokens. 

\mypara{Basic insight} \emph{When an LLM processes a context in a single pass, some tokens may not receive sufficient attention. By prompting the LLM to read the context multiple times, each with a different prefix prompt and thus a different KV cache, we hypothesize that tokens with consistently high attention scores across these KV caches are the ones requiring more attention.}

\begin{figure}[H]
    \centering
\includegraphics[width=0.8\columnwidth]{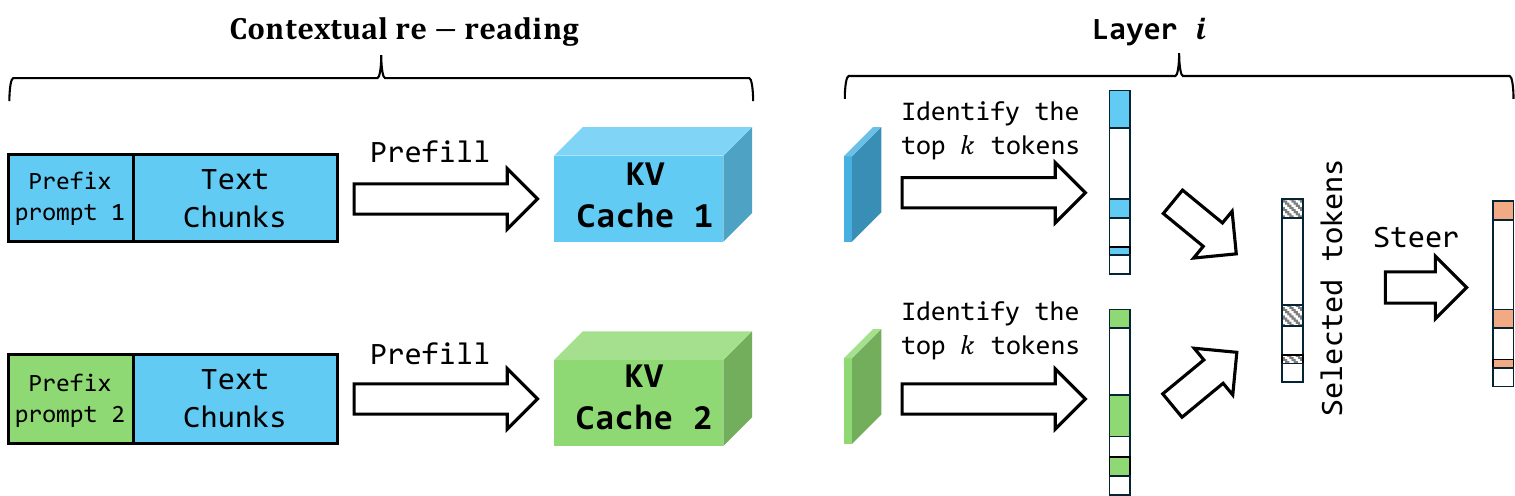} 
    \vspace{-0pt}
    \caption{\name system in an LLM inference workflow processes the same context twice using different prefix prompts to generate key-value caches. It then reweights the selected tokens.}
    \vspace{-5mm}
    \label{fig:workflow}
\end{figure}

\mypara{Key components} \name consists of three high-level steps. First, it prompts the LLM to process the context text chunks twice, each time using a different prefix prompt. This approach leverages the insight that varying the prefix prompt influences how the LLM interprets the context. Second, at each layer, it computes the cumulative attention scores for each head, selecting the top $k$ tokens based on these scores across all heads. It then identifies tokens that rank highly in both the first and second passes. Finally, it scales up the attention weight of selected tokens. An illustrative overview of \name is shown in Figure \ref{fig:workflow} and the corresponding algorithm can be found in \S\ref{subsec:appendix:algo}.


\mypara{Contextual re-reading} In this step, \name prompts the LLM to process the same context twice, each time using a distinct prefix prompt. The goal is to force the model to read the context in two different ways. The prefix prompts do not include the query itself because we want to ensure that the context is read in a non-query-dependent manner. This allows the model to capture general context-related information without bias toward a specific query. More importantly, given a fixed context $C$ and a sequence of queries, our method makes the prefix cache reusable across all queries and reduces the need to call the LLM twice for each query during online inference.


\mypara{Token selection} For each pass, we store the cumulative attention scores of the prefix prompt and context tokens at each layer and each head. For simplicity, we will ignore indices for layers and heads. Specifically, for each layer and each head, consider the input prompt tensor $\mathbf{X} \in \mathbb{R}^{L \times D}$, where $L$ denotes the input prompt length and $D$ represents the model’s hidden dimension. The attention matrix is computed as:
\[
    \mathbf{A} = softmax\left(\frac{\mathbf{Q} \mathbf{K}^{\top}}{\sqrt{D}}\right) \in \mathbb{R}^{L \times L},
\]
where $\mathbf{Q} = \mathbf{X} \mathbf{W}^Q$ and $\mathbf{K} = \mathbf{X} \mathbf{W}^K$. Here, $\mathbf{W}^Q, \mathbf{W}^K \in \mathbb{R}^{D \times D}$ are learned weight matrices that project the input prompt $\mathbf{X}$ into query and key spaces. Let $L_p$ denote the combined length of the prefix prompt and context. The attention matrix for the prefix prompt and context tokens is:
\[
    \mathbf{A}_p = \mathbf{A}[:L_p,:L_p] \in \mathbb{R}^{L_p \times L_p}.
\]
To compute the cumulative attention scores, we sum the attention weights across each column of the attention matrix. This summation gives us the total attention that each token has received from all preceding tokens:
\[
    \mathbf{A}_s = \sum_{i=0}^{L_p-1} \mathbf{A}_p[i, :] \in \mathbb{R}^{L_p},
\]
where $\mathbf{A}_{s}(i)$ is the cumulative attention score for token $i$, representing the total attention received by token $i$ from all preceding tokens.

We then store the cumulative attention scores from all heads and layers along with their corresponding indices for further processing. To determine which tokens to steer, we first sort the cumulative attention scores for each layer across all heads and identify the top $k$ tokens for that layer. Next, we find the intersection of these top-ranked tokens from both passes, selecting those that consistently rank high in both. These selected tokens are then upweighted.

\mypara{Steering} In the steering phase, for each layer, \name adjusts the attention mechanism by applying a weighting matrix, denoted as $\boldsymbol{M}$, which scales the attention scores for the selected tokens. Initially, we define $\boldsymbol{M} \in \mathbb{R}^{L \times L}$, where $L$ is the input prompt length, as a tensor filled with ones. For the selected tokens at each layer, we update the corresponding entries in $\boldsymbol{M}$ using a predefined scaling factor, $\alpha$. We then expand $\boldsymbol{M}$ across the batch and head dimensions to align with the shape of the attention weights, creating a tensor of shape $\mathbb{R}^{1 \times H \times L \times L}$, where $H$ represents the number of attention heads. Finally, we apply this weighting matrix element-wise to the original attention scores at each layer, effectively steering the attention toward the selected tokens:
\[
    \mathbf{A}_{\text{steered}} = \boldsymbol{M} \odot \mathbf{A},
\]
where $\odot$ denotes element-wise multiplication and $\mathbf{A}$ represents the original attention matrix.

%% file: sec_eval.tex
\section{Evaluation}
\vspace{-2mm}
\subsection{Setup}
\label{subsec:eval-setup}
\vspace{-2mm}
\mypara{Datasets} 
Our evaluation covers three datasets: SQuAD (~\cite{rajpurkar2016squad100000questionsmachine}), TriviaQA (~\cite{joshi2017triviaqalargescaledistantly}) and GSM8K (~\cite{cobbe2021trainingverifierssolvemath}).
SQuAD is a reading comprehension dataset with questions based on Wikipedia articles. TriviaQA is a large reading comprehension dataset, requiring reasoning across multiple sentences. GSM8K is a collection of diverse grade school math word problems designed for multi-step reasoning. For each dataset, we randomly sample 100 test cases. We use F1-score as the quality metric and request delay (in seconds) as the efficiency metric across all datasets.

\mypara{\name setting} We apply \name to Llama-8B, using an attention scaling factor $\alpha$. The two prefix prompts we used to generate different KV cache are provided in \S\ref{subsec:appendix:prefixes}. Following the AutoPASTA paper, we adopt coarse-to-fine model profiling on \name to search for the optimal set of layers and heads to steer that yields the most quality improvement.

\mypara{Baselines} We compare \name with three baselines: Llama-8B, Llama-70B (8-bit quantized) and Llama-8B with AutoPASTA. We also adopt coarse-to-fine model profiling on AutoPASTA. We assume prefix caching is enabled. That is, the KV cache of the prefix instruction and the context is pre-computed and has already been loaded to GPU memory when the request arrives.

\mypara{Implementation} We build \name and all the baselines on top of Huggingface Transformers (~\cite{transformers-library}). We implement AutoPASTA based on their paper as the code is not public yet. All experiments are conducted on two NVIDIA A40 GPUs.

\subsection{Results}
\begin{figure}[t]
    \centering
    \includegraphics[width=1.0\columnwidth]{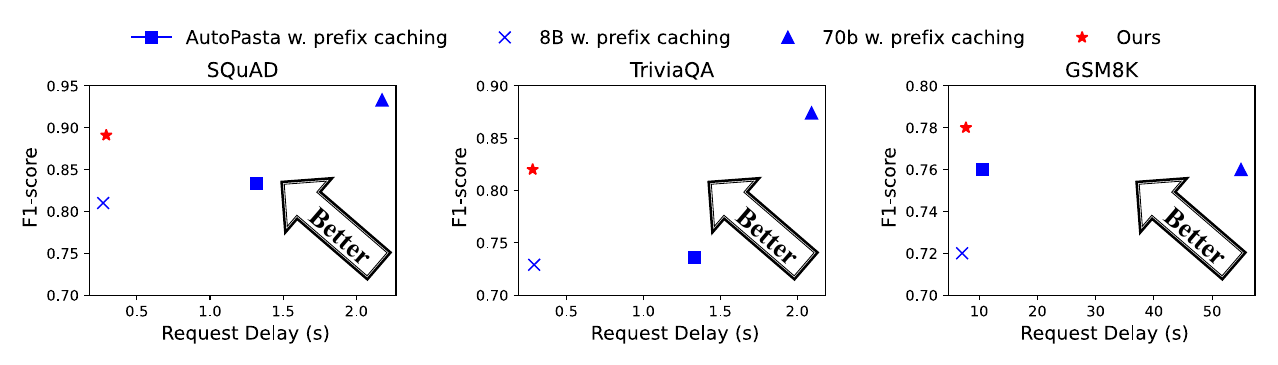}
    \vspace{-15pt}
    \caption{End-to-end Delay vs. Generation Quality. We assume KV cache is already in GPU memory when the request is being served.}
    \vspace{-5mm}
    \label{fig:overall_results}
\end{figure}

\vspace{-2mm}
\mypara{Reduced request delay} As shown in Figure \ref{fig:overall_results},  \name consistently achieves request delays close to the 8B baseline across all datasets. On the SQuAD, TriviaQA, and GSM8K datasets, \name's request delay is nearly negligible compared to the 8B baseline. Compared to the 70B model, \name is able to reduce request delay by 7.1–7.5×. Compared to AutoPASTA, \name can still reduce the request delay by 1.4–4.8×.

\mypara{Higher quality} Figure \ref{fig:overall_results} demonstrates the high generation quality of \name across all three datasets. On the SQuAD dataset, \name increases the F1 score by approximately 10\% over the LLaMA-8B model with prefix caching and by about 7\% over AutoPASTA. This improvement continues on the TriviaQA dataset, where \name achieves a similar F1 score increase, surpassing the 8B model by 12.5\% and AutoPASTA by 11.4\%. While our method does not outperform the 70B model on the SQuAD and TriviaQA datasets,  it closely approaches the performance of the larger model without the substantial delay associated with the larger model. Notably, on the GSM8K dataset, our method even exceeds the performance of the 70B model.

\mypara{Understanding \name's improvement} Unlike AutoPASTA, \name steers attention without relying on query information, operating with less information yet achieving superior performance. We test a query-dependent version of \name and observe smaller improvements compared to our query-independent method, suggesting that effective attention steering can be achieved without incorporating query information.

%% file: sec_future.tex
\section{Limitations and Future work}
\label{sec:limitation}
\vspace{-2mm}
We plan to extend to longer context lengths (e.g., >10k tokens) to fully explore \name's capabilities and to test it on models beyond Llama-8B to assess generalizability. We will also conducting an ablation study to quantify the contributions of the steering mechanism versus contextual re-reading. We will explore the limitations and capabilities of our method compared to traditional fine-tuning, especially considering context window length and model reasoning capabilities. Additionally, PagedAttention (\cite{vllm}) and FlashAttention (\cite{dao2022flashattention}) could enhance efficiency. Investigating more fine-grained steering at the granularity of individual token pairs, beyond our current token-level attention upweighting, may further improve generation quality and will be explored in future work.

%% file: sec_ending.tex
\section{Conclusion}
\label{sec:ending}
\vspace{-2mm}
We introduce \name, a novel attention steering method aimed to improve the generation quality of LLMs by allowing the model to review the context multiple times. Specifically, \name narrows the response quality gap between small and large LLMs by 65.9\% and deliver the responses faster by 4.8× compared to the state-of-the-art attention steering baseline.